\def\year{2021}\relax
\documentclass[letterpaper]{article} 
\usepackage{aaai21}  
\usepackage{times}  
\usepackage{helvet} 
\usepackage{courier}  
\usepackage[hyphens]{url}  
\usepackage{graphicx} 
\usepackage{natbib}  
\usepackage{caption} 
\usepackage{amsmath} 
\usepackage{amssymb}  
\usepackage{amsthm}
\usepackage{color,soul} 
\usepackage[ruled,vlined,linesnumbered]{algorithm2e}
\usepackage{mathtools}
\usepackage{graphics} 
\usepackage{subcaption}

\title{Ensuring Monotonic Policy Improvement \\ in Entropy-regularized Value-based Reinforcement Learning}
\author{Lingwei Zhu and Takamitsu Matsubara\\ 
}
\affiliations{

    Graduate School of Science and Technology, Nara Institute of Science and Technology, Japan\\
    zhu.lingwei.zj5@is.naist.jp, takam-m@is.naist.jp

}

\begin{document}

\maketitle

\begin{abstract}
This paper aims to establish an entropy-regularized value-based reinforcement learning method that can ensure the monotonic improvement of policies at each policy update. Unlike previously proposed lower-bounds on policy improvement in general infinite-horizon MDPs, we derive an entropy-regularization aware lower bound. Since our bound only requires the expected policy advantage function to be estimated, it is scalable to large-scale (continuous) state-space problems.   We propose a novel reinforcement learning algorithm that exploits this lower-bound as a criterion for adjusting the degree of a policy update for alleviating policy oscillation. We demonstrate the effectiveness of our approach in both discrete-state maze and continuous-state inverted
pendulum tasks using a linear function approximator for value estimation.

\end{abstract}

\section{Introduction}

\noindent Reinforcement Learning (RL) \cite{Sutton-RL2018} has recently achieved impressive successes in fields such as robotic manipulation \cite{openai2019solving}, video game playing \cite{mnih2015human} and the game of Go \cite{Silver2016}. However, compared with supervised learning that has wide-range of practical applications, RL applications have primarily been limited to casual game playing or laboratory based robotics. A crucial reason for limiting applications to these environments is that it is not guaranteed that the performance of RL policies will improve monotonically; they often oscillate during policy updates. As such, deploying such updated policies without examining its reliability might bring severe consequences in real-world scenarios, e.g., crashing a self-driving car.

Dynamic programming (DP) \cite{Bertsekas2005} offers a well-studied framework under which strict policy improvement is possible: with known state transition model, reward function and exact computation, monotonic improvement is ensured, and convergence is guaranteed within a finite number of iterations \cite{Ye2011}. However, in practice an accurate model of the environment is rarely available. In situations where the either model knowledge is absent, or the DP value functions cannot be explicitly computed, approximate DP and corresponding RL methods are to be considered. However, approximation introduces unavoidable update and Monte-Carlo sampling errors, and possibly restricts the policy space in which the policy is updated, leading to \emph{policy oscillation} phenomenon \cite{Bertsekas2011,wagner2011}, whereby the updated policy performs worse than pre-update policies during intermediary stages of learning. Inferior updated policies resulting from policy oscillation might pose a physical threat to real-world RL applications. Further, as value-based methods are widely employed in the state-of-the-art RL algorithms \cite{BHATNAGAR2009,haarnoja-SAC2018}, addressing the problem of policy oscillation becomes imminent.

Previous studies \cite{Kakade02,pirotta13} have attempted to address this issue by deriving lower bounds of policy improvement that evaluate the quality of updated policies. However, estimating such lower-bounds are intractable for practical RL scenarios except for small problems \cite{pirotta13} due to their complexity. 
A significant factor causing the complexity might be its excessive generality \cite{Kakade02,pirotta13}; 
Those bounds do not focus on any particular class of value-based RL algorithms. 
In this paper, in order to develop more tractable bounds, we focus on an RL class known as entropy-regularized value-based methods \cite{azar2012dynamic,Fox2016,haarnoja-SAC2017a,haarnoja-SAC2018}, where the entropies of policies are introduced in the reward function for regularizing policy updates. Sample efficiency and error-tolerance have been well-studied \cite{kozunoCVI}; however, their monotonic improvement has not been explored.

In this paper, we aim to establish an entropy-regularized value-based reinforcement learning method that can ensure the monotonic improvement of policies. 
Unlike previously proposed lower-bounds on policy improvement in general infinite-horizon MDPs, we derive an entropy-regularization {\it aware} lower bound on policy improvement in the infinite-horizon entropy-regularized MDPs. Since our bound only requires the expected policy advantage function to be estimated, it is scalable to large-scale (continuous) state-space problems.   We propose a novel reinforcement learning algorithm that exploits this lower-bound as a criterion for adjusting the degree of a policy update for alleviating policy oscillation. We demonstrate the effectiveness of our approach in both discrete-state maze and continuous-state inverted pendulum tasks using a linear function approximator for value estimation.




The remainder of this paper is organized as follows. After a brief review on related work, we provide a preliminary on RL and proceed to the theory of the proposed algorithm. Experimental results are followed by discussions and conclusion. All proofs are deferred until appendix.

\section{Related Work}\label{S2}

The policy oscillation phenomenon, also termed \emph{overshooting} by \cite{wagner2011}, referred to as degraded performance of updated policies, frequently arises in approximate policy iteration algorithms \cite{Bertsekas2011}, and can occur even under asymptotically converged value functions \cite{wagner2011}. It has been shown that aggressive updates with sampling and update errors, together with restricted policy spaces, are the main reasons for policy oscillations \cite{pirotta13}. 

To attenuate policy oscillation, \cite{Kakade02} proposed Conservative Policy Iteration (CPI) whereby the greedily updated policy is interpolated with the current policy to achieve less aggressive updates. 
Several algorithms were proposed by \cite{pirotta13,abbasi-improvement16} to improve upon CPI by proposing new lower bounds for policy improvement. However, since their focus is general stationary policies, deriving practical algorithms based on the lower bounds is nontrivial. This might explain why heuristics must be added in \cite{Vieillard-2020DCPI} to extend CPI to be compatible with neural networks. To remove this limitation, our focus on entropy-regularized policies allows for a straightforward algorithm based on a novel, significantly simplified lower bound. 

Another line of research to alleviating policy oscillation is to exploit the idea of \emph{gap-increasing} operators \cite{azar2012dynamic,Bellemare2016,asadi17a,kozunoCVI}. Instead of interpolating greedy policies, smaller updates are taken in the stochastic policy space by enforcing similarity (e.g. relative entropy) between updates. By incorporating entropy terms, entropy-regularized methods have recently achieved impressive successes \cite{haarnoja-SAC2017a,haarnoja-SAC2018,ZHU2020CEP}. However, those algorithms do not explicitly consider ensuring monotonic improvement as we do in this paper.

The concept of monotonic improvement has been exploited also in policy search scenarios \cite{trpo-schulman15,akrour-monotonic-2016}. It is worth noting that though \cite{trpo-schulman15,akrour-monotonic-2016} have demonstrated good empirical performance, they focus on local optimal policy with strong dependency on initial parameters. On the other hand, we focus on value-based RL which searches for global optimal policies.

\section{Preliminary}\label{S3}


\subsection{Value-based Reinforcement Learning}

RL problems can be formulated by Markov Decision Processes (MDPs) expressed by the quintuple $(\mathcal{S},\mathcal{A},\mathcal{T},\mathcal{R},\gamma)$, where $\mathcal{S}$ denotes the state space, $\mathcal{A}$ denotes the finite action space, $\mathcal{T}$ denotes transition dynamics such that $\mathcal{T}_{ss'}^{a}\!:=\!\mathcal{T}(s'|s,a)$ represents the transition from state $s$ to $s'$ with action $a$ taken. $\mathcal{R} \!=\! r^{\,\,a}_{ss'}$ is the immediate reward associated with that transition. In this paper, we consider $r^{\,\,a}_{ss'}$ being bounded in the interval $[-1, 1]$.  $\gamma \in (0,1)$ is the discount factor. For simplicity we consider the infinite horizon discounted setting with a fixed starting state $s_{0}$. A policy $\pi$ is a probability distribution over actions given some state. We also define the stationary state distribution induced by $\pi$ as $d^{\pi}(s) \!=\! (1-\gamma)\!\sum_{t=0}^{\infty}\gamma^{t}\mathcal{T}({s_{t} \!=\! s|s_{0}, \pi})$. Throughout this paper, for all notations $(\cdot)^{\alpha}_{\beta, d}$ with $\alpha$ and $\beta$ being policies, $d$ refers to $d^{\alpha}$.

RL methods search for an optimal stationary policy $\pi^{*}$ such that the expected long-term discounted reward is maximized, over all states:
\begin{align}
  \begin{split}
    V_{\pi^{*}}(s) = \max_{\pi}\mathbb{E}_{\mathcal{T}}\big[ \sum_{t=0}^{\infty}\gamma^{t} (r_{ss'}^{a})_{t} \big| s_{0} = s \big].
  \end{split}
\end{align}
It is known that $V_{\pi^{*}}$ solves the following system of equations known as the Bellman optimality \cite{Sutton-RL2018}:
\begin{align}
\begin{split}
V_{{\pi}^{*}}(s)=\max_{\pi}{\sum_{\substack{a \in \mathcal{A} \\ s' \in \mathcal{S}}}\pi(a|s)\bigg[\mathcal{T}_{ss'}^{a}\big(r_{ss'}^{a}+\gamma V_{{\pi}^{*}}(s')\big)\bigg]}.
\label{sys_Vbellman}
\end{split}
\end{align}
The state-action value function $Q_{\pi^{*}}(s,a)$ is more frequently used in control context:
\begin{align}
\begin{split}
Q_{\pi^{*}}(s,a)&=\max_{\pi}{\sum_{s'\in\mathcal{S}}{\mathcal{T}_{ss'}^{a}\big(r_{ss'}^{a}\!\!+\!\gamma\!\sum_{a'\in\mathcal{A}}{\pi(a'|s')Q_{\pi^{*}}(s',a')}\big)}}.
\label{sys_Qbellman}
\end{split}
\end{align}

\subsection{Lower Bounds on Policy Improvement}

The following lemma formally defines the criterion of policy improvement of policy $\pi'$ over $\pi$:

\textbf{Lemma 1} \cite{Kakade02}. \emph{For any stationary policies $\pi'$ and $\pi$ the following equation holds}:
\begin{align}
  \begin{split}
    &\Delta J^{\pi'}_{\pi, d} \!:=\! J^{{\pi'}}_{d} \!-\! J^{{\pi}}_{d} \!=\!\!  {\sum_{s}{d^{\pi'}(s)\!\sum_{a}{\!\pi'(a|s){A_{\pi}(s,a)}}}}, \\
    &\text{where } J^{{\pi'}}_{d} \!:=\! \mathbb{E}_{s_{0},a_{0},\dots}{\bigg[\sum_{t=0}^{\infty}\gamma^{t}r_{t}\bigg]} \!=\! \sum_{s}{d^{{\pi'}}\!\!(s)}\!\sum_{a}{{\pi'}(a|s)r^{a}_{ss'}}.\\    
  \end{split}
  \label{kakade_idty}
\end{align}
where $A_{\pi}(s,a) \!:=\! Q_{\pi}(s,a) - V_{\pi}(s)$ is the advantage function. Though Lemma 1 relates policy improvement to the expected advantage function, pursuing policy improvement by directly exploiting Lemma 1 is intractable as it requires comparing $\pi'$ and $\pi$ point-wise for infinitely many new policies. Many existing works \cite{Kakade02,pirotta13,trpo-schulman15} instead focus on finding a $\pi'$ such that the right-hand-side of Eq. (\ref{kakade_idty}) is lower-bounded. In order to alleviate policy oscillation brought by the greedily updated policy $\tilde{\pi}$, \cite{Kakade02} proposes to adopt \emph{partial update}:
\begin{align}
  \begin{split}
\pi' = \zeta\tilde{\pi} + (1-\zeta)\pi.
  \label{mixture_policy}
  \end{split}
\end{align}
to interpolate between the greedy policy and the current policy to achieve conservative updates. 

Following this concept, \cite{pirotta13} proposes to optimize the coefficient $\zeta$ to attain a maximum lower bound $\Delta J^{\pi'}_{\pi, d}$ on policy improvement. The optimal value $\zeta^{*}$ hence represents the optimal policy of a linear policy class spanned by $\tilde{\pi}$ and $\pi$. The following lemma relates the lower bound on improvement to the maximum total variation of $\tilde{\pi}, \pi$:

\textbf{Lemma 2} \cite{pirotta13} \emph{Provided that [1] policy $\pi'$ is generated by partial update Eq. (\ref{mixture_policy}); [2] $\zeta$ is chosen properly and [3] $A_{\pi, d}^{\tilde{\pi}} \geq 0$, then the following improvement is guaranteed:} 
\begin{align}
  \begin{split}
&\Delta J^{\pi'}_{\pi, d} \geq \frac{\big((1-\gamma)A_{\pi,d}^{\tilde{\pi}}\big)^{2}}{2\gamma\delta\Delta A^{\tilde{\pi}}_{\pi}}, \\
\text{with } & \zeta = \min{(1, \zeta^{*})},\\
\text{where } &\zeta^{*}=\frac{(1-\gamma)^{2}A^{\tilde{\pi}}_{{\pi, d}}}{\gamma\delta\Delta A^{\tilde{\pi}}_{\pi}},\\
&\delta=\max_{s}{\big|\!\sum_{a\in\mathcal{A}}\big(\tilde{\pi}(a|s)-\pi(a|s)\big)\big|},\\
&\Delta A^{\tilde{\pi}}_{\pi}=\max_{s, s'}{|A^{\tilde{\pi}}_{\pi}(s)-A^{\tilde{\pi}}_{\pi}(s')}|,
  \end{split}
  \label{J_first_exact}
\end{align}
where $A^{\tilde{\pi}}_{\pi, d}\!:=\!\sum_{s}d^{\pi'}\!(s) A^{\tilde{\pi}}_{\pi}\!(s)$ is the expected policy advantage and $A^{\tilde{\pi}}_{\pi}\!(s) \!=\! \sum_{a}\big(\tilde{\pi}(a|s) \!-\! \pi(s,a)\big)Q_{\pi}(s,a)$ is the policy advantage function.

\begin{proof}
 See the Proof of Lemma 2 in Appendix.
\end{proof}

By noting that $\tilde{\pi}(a|s) - \pi(s,a)$ appears in both $\delta$ and $\Delta A^{\tilde{\pi}}_{\pi}$, we see that the policy improvement $\Delta J^{\pi'}_{\pi, d}$ is governed by the maximum total variation of policies. While one can exploit Lemma 2 for a value-based RL algorithm, it is obvious that it could only apply to problems with small state-action spaces. In general, without further assumptions on $\pi', \tilde{\pi}, \pi$, lower bounding policy improvements is intractable, as maximization $\delta$ and $\Delta A^{\tilde{\pi}}_{\pi}$ in large state space require exponentially many samples for accurate estimation.

In the next section, we propose a novel lower bound on policy improvement and a scalable algorithm applicable to large state spaces by exploiting entropy-regularized policies.

\section{Proposed Method}\label{S44}

In this section we detail our proposed method. First a general formulation of entropy-regularized RL is introduced, followed by a lemma that bounds the maximum distance between policies of entropy-regularized update. Finally we propose the main theorem and a novel algorithm for ensuring monotonic improvement.

\subsection{Entropy-regularized RL}

We provide a general formulation for entropy-regularized algorithms \cite{azar2012dynamic,haarnoja-SAC2018,kozunoCVI} in the following. At iteration \emph{k}, the entropy of current policy $\pi_{k}$ and Kullback-Leibler (KL) divergence between $\pi_{k}$ and some baseline policy $\bar{\pi}$ are added to the value function:
\begin{align}
\begin{split}
V_{\bar{\pi}}^{\pi_{k}}(s)\!&=\!\sum_{\substack{a \in \mathcal{A} \\ s' \in \mathcal{S}}}\!\pi(a|s)\!\bigg[\mathcal{T}_{ss'}^{a}\big(r_{ss'}^{a}+\gamma V^{*}_{\bar{\pi}}(s')\big) \!-\! \mathcal{I}_{\bar{\pi}}^{\pi_{k}}\!\bigg],\!\! \\
\mathcal{I}_{\bar{\pi}}^{\pi_{k}} &= -\tau\log{\pi_{k}(a|s)} - \sigma\log{\frac{\pi_{k}(a|s)}{\bar{\pi}(a|s)}},
\label{sys_DPPbellman}
\end{split}
\end{align}
where $\tau$ controls the weight of entropy bonus and $\sigma$ weights the effect of KL regularization. The baseline policy $\bar{\pi}$ is often taken as the previous iteration policy $\pi_{k-1}$. For notational convenience, in the remainder of this paper, we define $\alpha := \frac{\tau}{\tau+\sigma}$, $\beta:=\frac{1}{\tau+\sigma}$. Intuitively, the entropy term enables multi-modal policy behavior \cite{haarnoja-SAC2017a} and the KL divergence provides smooth policy updates \cite{azar2012dynamic,kozunoCVI}. When the optimal policy is attained, the KL regularization term is zero. Hence the optimal policy maximizes the cumulative reward while keeping the entropy high.

It is worth noting that several upper bounds of the form $|\!|J^{*}-J^{k}|\!|_{\infty}$ for entropy-regularized RL exist \cite{azar2012dynamic,kozunoCVI}. However, we are unaware of any general lower bound that guarantees monotonic improvement like $|\!|J^{k+1}-J^{k}|\!| \geq 0$ for entropy-regularized algorithms.

\subsection{Bounding Policy Update by Entropy Regularization}

The core concept of Lemma 2 is lower-bounding policy improvement by upper-bounding the stationary distribution difference $d^{\pi'}\!-\!d^{\pi}$ with maximum total variation $\delta$ \cite{pirotta13}. However, besides the assumption of stationarity, it is intractable to solve $\delta$ over large state spaces without further specification on the considered policy class. 

Our approach is based on the aforementioned entropy-regularized value-based algorithms that have achieved state-of-the-art performance on several benchmark problems \cite{haarnoja-SAC2018,ZHU2020CEP}. A very recent study of which offers a means to bound the maximum distance between pre- and post-update policies \cite{kozunoCVI}. The key insight of our approach is that by considering the class of entropy-regularized policies, Lemma 2 can be significantly simplified to apply to large state spaces. To begin with our derivation, we first introduce the following lemma:


\textbf{Lemma 3} \cite{kozunoCVI}. \emph{For any entropy-regularized policies $\pi_{k}$ and $\pi_{k+1}$ generated by value functions Eq. (\ref{sys_DPPbellman}), the following bound holds for their maximum KL divergence}:

\begin{align}
  \begin{split}
    &\max_{s}{D_{KL}\big(\pi_{k+1}(\cdot|s) |\!| \pi_{k}(\cdot|s)\big)} \leq 4 B_{k} + 2 C_{k},\\
    \text{where } &B_{K}=\frac{1-\gamma^{K}}{1-\gamma}\epsilon\beta , \,\, C_{K} = \beta r_{max} \sum_{k=0}^{K-1}{\alpha^{k}\gamma^{K-k-1}},
  \end{split}
  \label{CVI_kl}
\end{align}
\emph{where $K$ and $k$ are any positive integers, $\epsilon$ is the uniform upper bound of error.} 

\begin{proof}
  See appendix C.3 of \cite{kozunoCVI}.  
\end{proof}

Since the reward is bounded in $[-1, 1]$, $r_{max}$ can be conveniently dropped. Also for simplicity, in this paper we assume there is no update error, i.e., $B_{K}=0$. However, it is straightforward to extend to cases where errors present. 
Intuitively, Lemma 3 ensures that an updated entropy-regularized policy will not deviate much from the previous policy. 

\subsection{Entropy-regularization Aware Lower Bound on Policy Improvement in Entropy-regularized MDPs}\label{nlb}

 Our aim is to ensure monotonic policy improvement given policy $\pi_{k}$ at any iteration $k$.
Following \cite{Kakade02,pirotta13}, we propose to construct a new monotonically improving policy as:
\begin{align}
  \begin{split}
    \tilde{\pi}_{k+1} = \zeta\pi_{k+1} + (1-\zeta)\pi_{k}.
  \end{split}
  \label{mixture_cvi}
\end{align}
It is now clear by comparing Eq. (\ref{mixture_policy}) with Eq. (\ref{mixture_cvi}) our proposal takes $\pi', \tilde{\pi}, \pi$ as $\tilde{\pi}_{k+1}, \pi_{k+1}, \pi_{k}$, respectively. It is worth noting that ${\pi}_{k+1}$ is the updated policy that has not been \emph{accepted} for deployment. 

Intuitively, the agent collects samples and updates the policy to $\pi_{k+1}$. However, instead of directly deploying this policy, we interpolate it with $\pi_{k}$ by $\zeta$ to obtain $\tilde{\pi}_{k+1}$. As is shown in Theorem 4, $\tilde{\pi}_{k+1}$ is optimal in the sense of providing largest improvement among the linear class of policies spanned by $\pi_{k+1}$ and $\pi_{k}$, in contrast to the point-wise comparison in Eq. (\ref{kakade_idty}).

\textbf{Theorem 4.} \emph{Provided that [1] partial update Eq. (\ref{mixture_cvi}) is adopted; [2] $A^{{\pi_{k+1}}}_{\pi_{k},d} \geq 0$ and [3] $\zeta$ is chosen properly, then any entropy-regularized policies generated by Eq.(\ref{sys_DPPbellman}) guarantees the following improvement that depends only on $\alpha, \beta, \gamma \text{ and } A^{{\pi_{k+1}}}_{\pi_{k},d}$ after any policy update:}

\begin{align}
  \begin{split}
\Delta J^{\tilde{\pi}_{k+1}}_{\pi_{k},d}  &\geq \frac{\big(1-\gamma)^{3}(A_{\pi_{k},d}^{{\pi_{k+1}}})^{2}}{16\gamma C_{K}},\\
\text{with } \zeta &= \min{(1, \zeta^{*})}, \\
\text{where } \zeta^{*} &= \frac{(1-\gamma)^{3}A^{{\pi_{k+1}}}_{{\pi_{k},d}}}{2\gamma C_{K}},\\
C_{K} &= \beta\sum_{k=0}^{K-1}{\alpha^{k}\gamma^{K-k-1}}.\\
\label{J_first_improved}
  \end{split}
\end{align}

\begin{proof}
  See the Proof of Theorem 4 in Appendix.
\end{proof}


Theorem 4 is one of the main contributions of this paper, in which $A^{{\pi_{k+1}}}_{{\pi_{k}},d}$ is the only quantity that needs to be estimated. It is worth noting that $A^{{\pi_{k+1}}}_{{\pi_{k}}}\geq 0$ is a straightforward criterion that is naturally satisfied by greedy policy improvement of the policy iteration when computation is exact. To handle the case when it is negative caused by error or approximate computations, we implement an optional simple \emph{rejection} mechanism to reject this update, as will be detailed in the summary of algorithm. 

\subsection{Algorithm for Ensuring Monotonic Improvement}

We now detail the structure of our proposed algorithm based on Theorem 4. Specifically, value update, policy update and stationary distribution estimation are introduced, followed by a short discussion on update rejection.

\subsubsection{Value Update}

In order to estimate $A_{\pi_{k},d}^{{\pi_{k+1}}}$ in Theorem 4, both $A_{\pi_{k}}^{{\pi_{k+1}}}$ and $d^{\tilde{\pi}_{k+1}}$ need to be estimated from samples. Since $A_{\pi_{k}}^{{\pi_{k+1}}}(s) \!=\! \sum_{a}\pi_{k+1}(a|s)\big(Q_{\pi_{k}}(s,a) \!-\! V_{\pi_{k}}(s)\big)$, one needs an explicit form of $\pi_{k+1}(a|s)$. This step is general and may vary according to the algorithm used. For Mellowmax or Boltzmann policy \footnotemark $\pi_{k+1}(a|s) \!\sim\! \exp\big(\beta Q_{\pi_{k+1}}(s,a)\big)$ \cite{asadi17a,kozunoCVI}, we can first update the value functions using the empirical Bellman operator $\mathcal{B_{\pi}}$:
\begin{align}
    &Q_{\pi_{k+1}}(s,a) = \mathcal{B}_{\pi}Q_{\pi_{k}} := r_{ss'}^{a} + \gamma \sum_{a'}\pi_{k}(a'|s) Q_{\pi_{k}}(s', a'),
\label{empirical_bellman}
\end{align}
then evaluate the policy on the updated value function. 



\footnotetext{To be precise, Boltzmann policies in \cite{kozunoCVI} follow the form $\pi_{k} \sim \exp\big(\beta\Psi_{{k}}(s,a)\big)$, where $\Psi$ is an action preference function defined as $\Psi_{{k}}(s,a)=Q_{\pi_{k}}(s,a) -  \frac{\alpha}{\beta}\log\big(\pi_{k-1}(a|s) \big)$.}

\subsubsection{Policy Update}

The updated policy $\pi_{k+1}$ cannot be directly deployed since it has not been verified to improve upon $\pi_{k}$. We interpolate between $\pi_{k+1}$ and $\pi_{k}$ with coefficient $\zeta$ such that the resultant policy $\tilde{\pi}_{k+1}$ in Eq. (\ref{mixture_cvi}) achieves highest improvement $\Delta J^{\tilde{\pi}_{k+1}}_{\pi_{k},d}$ within the policy class spanned by $\pi_{k+1}$ and $\pi_{k}$.

Here, $\zeta$ is optimally tuned and dynamically changing in every update. It reflects the \emph{conservativeness} against policy oscillation, i.e., how much we trust the updated policy $\pi_{k+1}$. Generally, at the early stage of learning, $\zeta$ should be close to $0$ in order to explore conservatively.

\subsubsection{Estimating Stationary Distributions}

In practice, $d^{\tilde{\pi}_{k+1}}(s)$ in $A_{\pi_{k},d}^{{\pi_{k+1}}}$ of Eq. (\ref{J_first_improved}) is unwieldy as we have not deployed $\tilde{\pi}_{k+1}$. Motivated by \cite{Kakade02,trpo-schulman15}, we approximate Eq. (\ref{kakade_idty}) using 
\begin{align}
  \begin{split}
    \Delta {J^{\tilde{\pi}_{k+1}}_{\pi_{k}, d}} \approx \Delta \widehat{J^{\tilde{\pi}_{k+1}}_{\pi_{k}, d}} :=   {\sum_{s}{d^{\pi_{k}}(s) {\sum_{a} \tilde{\pi}_{k+1}(a|s) A_{\pi_{k}}(s)}}}, \\
  \end{split}
  \label{kakade_idty_approx}
\end{align}
where now stationary distribution $d^{\pi_{k}}$ is induced by $\pi_{k}$ instead of $\pi_{k+1}$ in $d^{\tilde{\pi}_{k+1}}$. If $\pi$ is a differentiable function parametrized by $\theta$, then according to \cite{Kakade02,trpo-schulman15}, $\Delta \widehat{J^{\tilde{\pi}_{k+1}}_{\pi_{k}, d}}$ and $\Delta {J^{\tilde{\pi}_{k+1}}_{\pi_{k}, d}}$ have the same first-order terms. 

Entropy-regularized policies further justify this approximation by allowing us to bound the maximum \emph{improvement loss} $\big| \Delta {J^{\tilde{\pi}_{k+1}}_{\pi_{k}, d}} - \Delta \widehat{J^{\tilde{\pi}_{k+1}}_{\pi_{k}, d}} \big|$ even without assuming $\pi$ is differentiable. We first prove the following novel result:

\textbf{Lemma 5. }\emph{For any entropy-regularized policies $\tilde{\pi}_{k+1}$ generated by Eq. (\ref{mixture_cvi}) and $\pi_{k+1}, \pi_{k}$ by Eq. (\ref{sys_DPPbellman}), the following holds:}
\begin{align*}
  &|\!| d^{\tilde{\pi}_{k+1}} - d^{\pi_{k}} |\!|_{1} \leq \frac{2\zeta\gamma}{(1-\gamma)^2}\sqrt{C_{K}}.
\end{align*}
\begin{proof}
  See the proof of Lemma 5 in Appendix. 
\end{proof}

Equipped with Lemma 5, we provide the following bound on the improvement loss induced by employing approximation Eq. (\ref{kakade_idty_approx}):

\textbf{Theorem 6. } \emph{Employing approximation Eq. (\ref{kakade_idty_approx}) for entropy-regularized policies can cause the improvement loss $\mathcal{L}^{a, d^{a}}_{b, d^{b}}$ of at most}:
\begin{align*}
  \big| \mathcal{L}^{a, d^{a}}_{b, d^{b}} \big| \!:=\! \bigg|\Delta {J^{\tilde{\pi}_{k+1}}_{\pi_{k}, d}} -  \Delta \widehat{J^{\tilde{\pi}_{k+1}}_{\pi_{k}, d}} \bigg| \leq (1-\gamma)|\!|A_{b}^{a}|\!|^{2}_{1} ,
\end{align*}
where $\tilde{\pi}_{k+1}$ is denoted as $a$ and $\pi_{k}$ as $b$.
\begin{proof}
  See the proof of Theorem 6 in Appendix.
\end{proof}
\textbf{Remark. } In practice $\gamma$ is typically set to a large value, e.g. $\gamma \!=\! 0.95$, then improvement loss $\mathcal{L}^{a, d^{a}}_{b, d^{b}} = 0.05 |\!|A_{b}^{a}|\!|^{2}_{1}$. When we bound the reward $r_{max} \!=\! 1$, it is often true that $|\!|A_{b}^{a}|\!|_{1} \!\leq\! 1$. Hence for entropy-regularized algorithms, changing the stationary distribution following Eq. (\ref{kakade_idty_approx}) can cause only small improvement loss.

\begin{algorithm}[t!]
  \SetAlgoLined
  \KwIn{$\gamma, \alpha, \beta, T, K, \pi_{0}$}
  \KwOut{learned policy $\tilde{\pi}^{*}$}
  \BlankLine
 $\zeta_{0} , Q_{\pi_{0}}, A_{\pi_{0}} = 0$
 \BlankLine
  \For{$k = 0, 1, \dots,K$}
  {
   \For{$t=1,\dots,T$ }
   {
     $s_{t+1}, r_{t+1}=\mathit{InteractAndObserve}(a_{t})$\\
     $D_{0:k} = \mathit{Collect}(s_{t}, a_{t}, r_{t}, s_{t+1}, r_{t+1})$\\
 }
 \BlankLine
 $Q_{\pi_{k+1}}, A_{\pi_{k+1}} \!=\! \mathit{ValueUpdate}(D_{1:k}, Q_{\pi_{k}}, A_{\pi_{k}})$\\
 \BlankLine
 $\pi_{k+1} = \mathit{PolicyUpdate}(D_{0:k}, Q_{\pi_{k+1}})$\\
 \BlankLine
 \For{every $s$}{
  $d^{\pi_{k}}(s), A^{{\pi_{k+1}}}_{\pi_{k}}(s) = \mathit{PolicyAdvantage}(A_{\pi_{k}},\pi_{k},\pi_{k+1})$\\
 }
 \BlankLine
 $A^{{\pi_{k+1}}}_{\pi_{k}, d} \!=\! \sum_{s} d^{\pi_{k}}(s) A^{{\pi_{k+1}}}_{\pi_{k}}(s)$\\
 \BlankLine

  \BlankLine
 \eIf{$A^{{\pi_{k+1}}}_{\pi_{k}, d}\geq 0$}{ 
   \BlankLine
   $\tilde{\pi}_{k+1} = \mathit{Interpolate}(\zeta,\pi_{k+1}, \pi_{k})$\\
 }{
 \emph{\# optional rejection}\\
    $\mathit{RejectUpdate}()$\\
}
 }
 \caption{Entropy-regularized Value-based Reinforcement Learning with Monotonic Improvement}
  \label{ARPU_alg}
 \end{algorithm}

\begin{figure*}[t!]
  \begin{subfigure}[]{0.24\textwidth}
    \includegraphics[width=\linewidth]{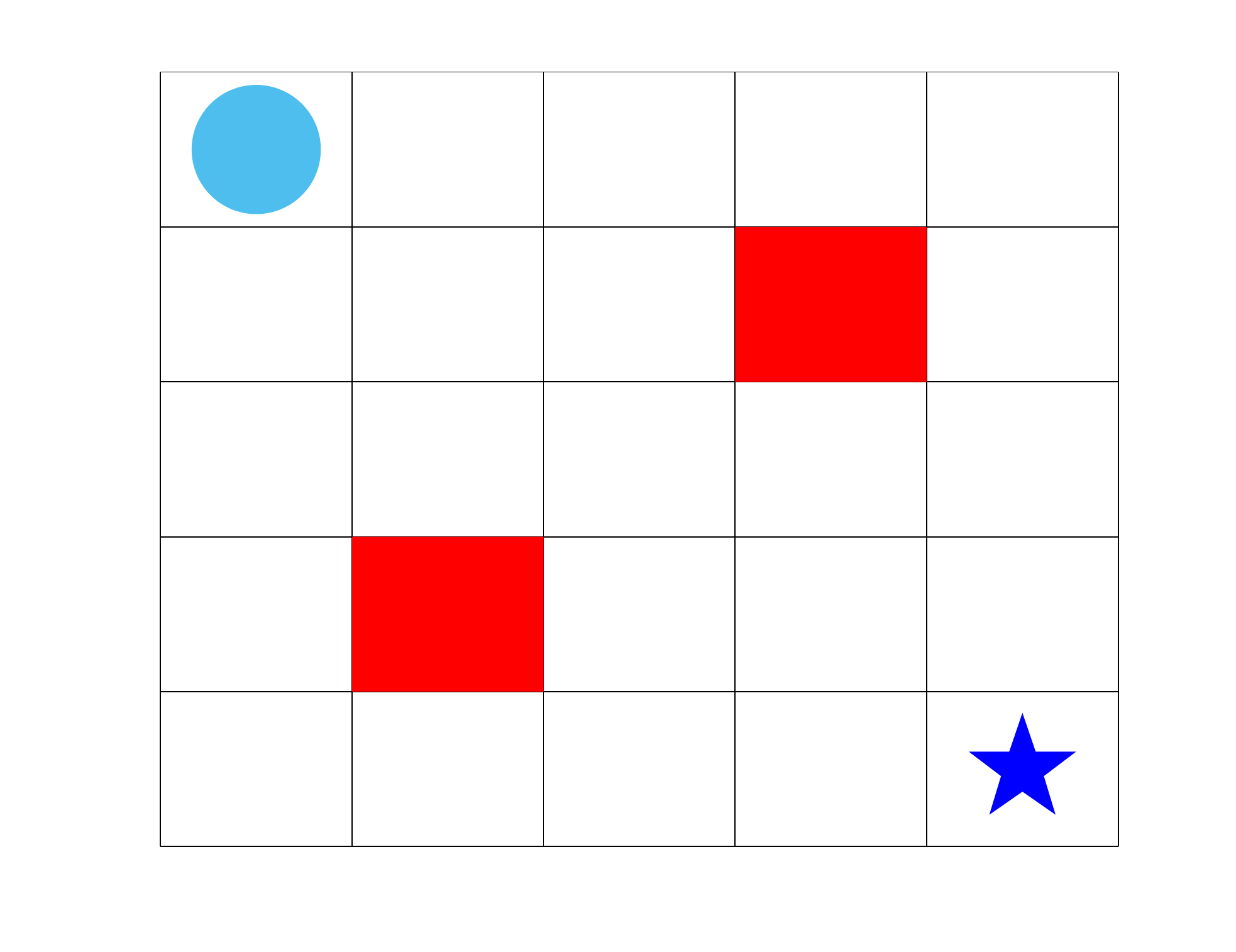}
    \caption{The $5\times 5$ gridworld}
    \label{gridworld}
  \end{subfigure}
  \begin{subfigure}[]{0.24\textwidth}
    \includegraphics[width=\linewidth]{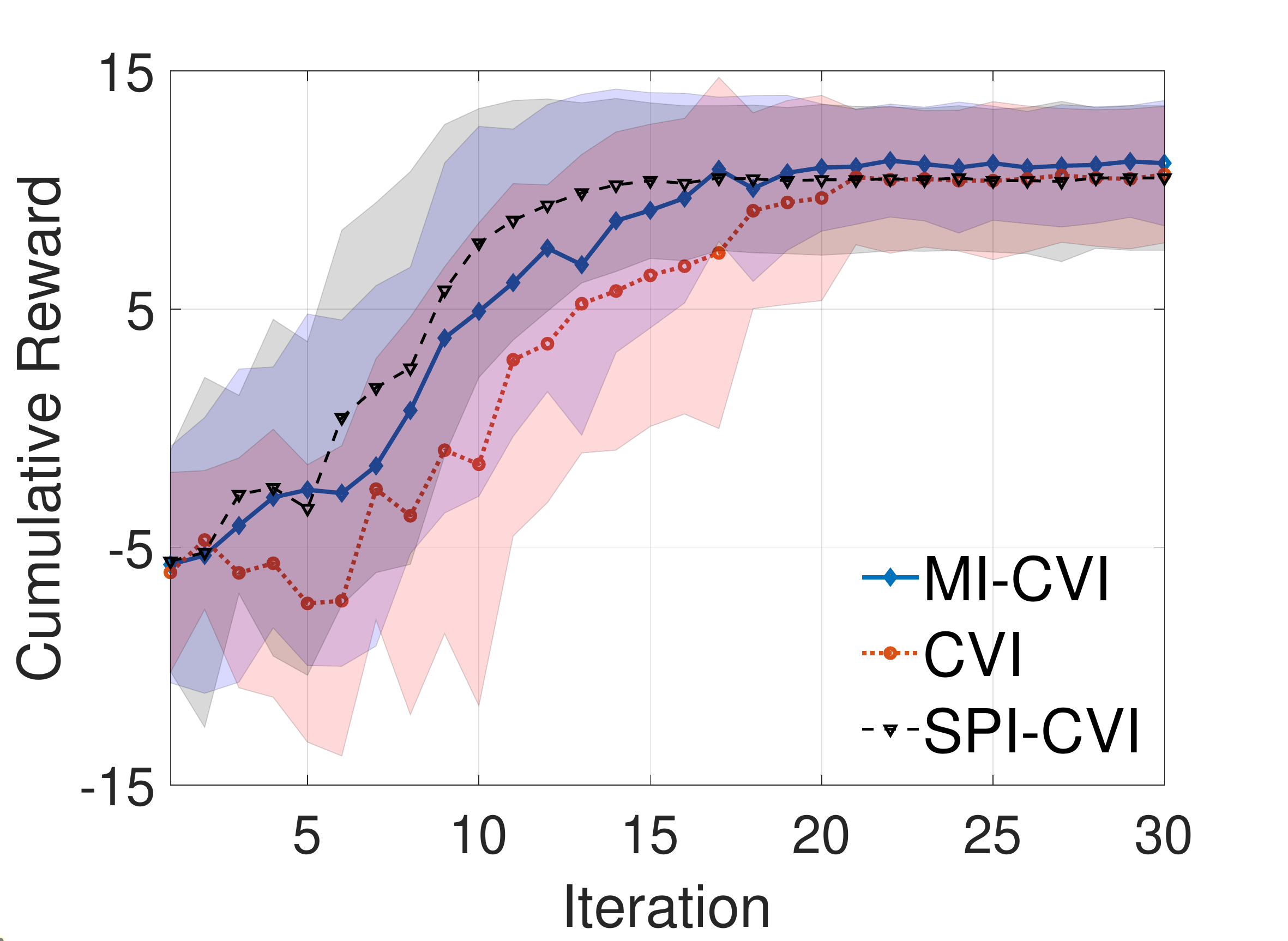}
    \caption{Cumulative reward}
    \label{comp_cr}
  \end{subfigure}
  \begin{subfigure}[]{0.24\textwidth}
    \includegraphics[width=\linewidth]{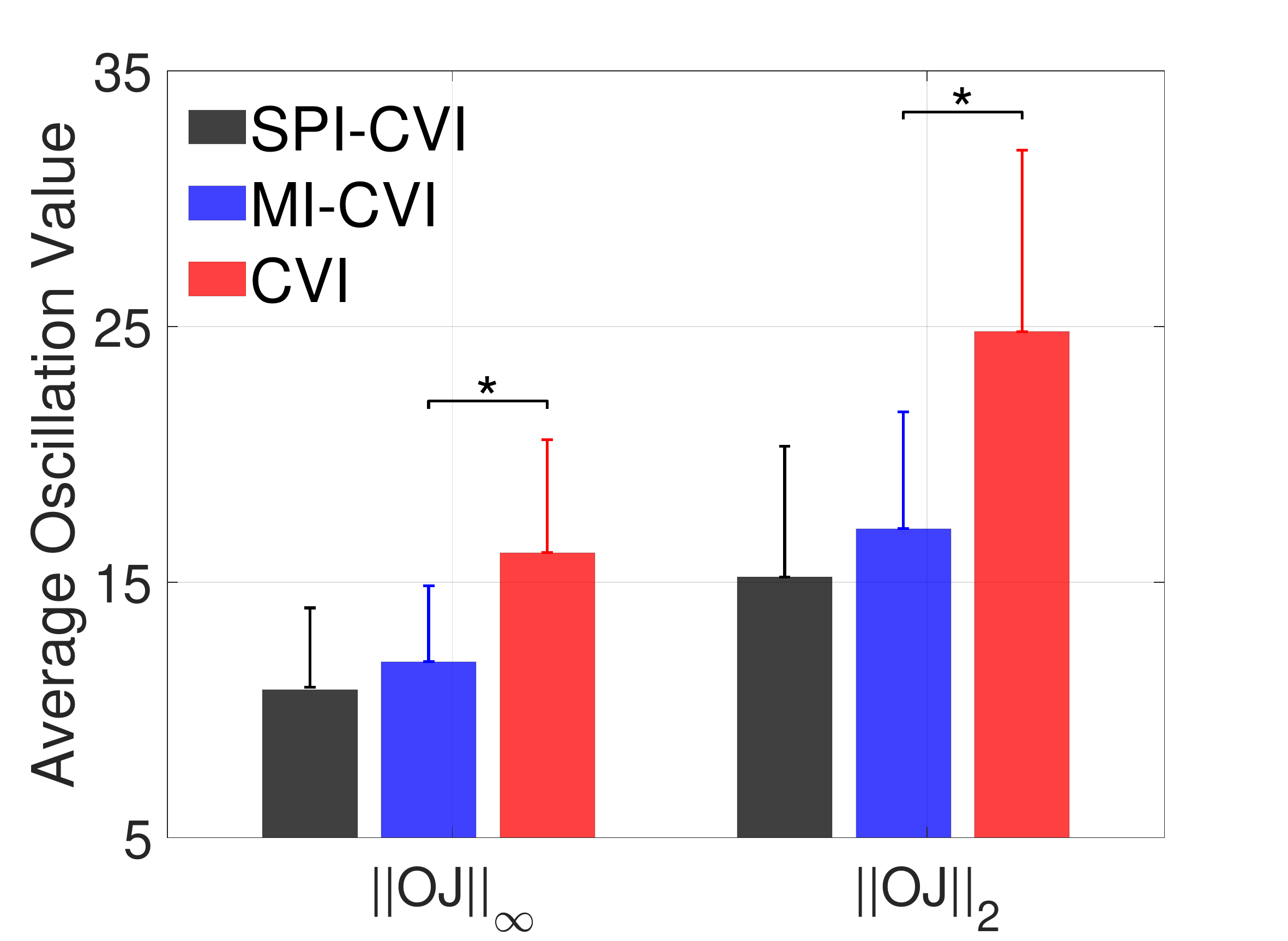}
    \caption{Values of policy oscillation}
    \label{gridworld_oscillation}
  \end{subfigure}
  \begin{subfigure}[]{0.24\textwidth}
    \includegraphics[width=\linewidth]{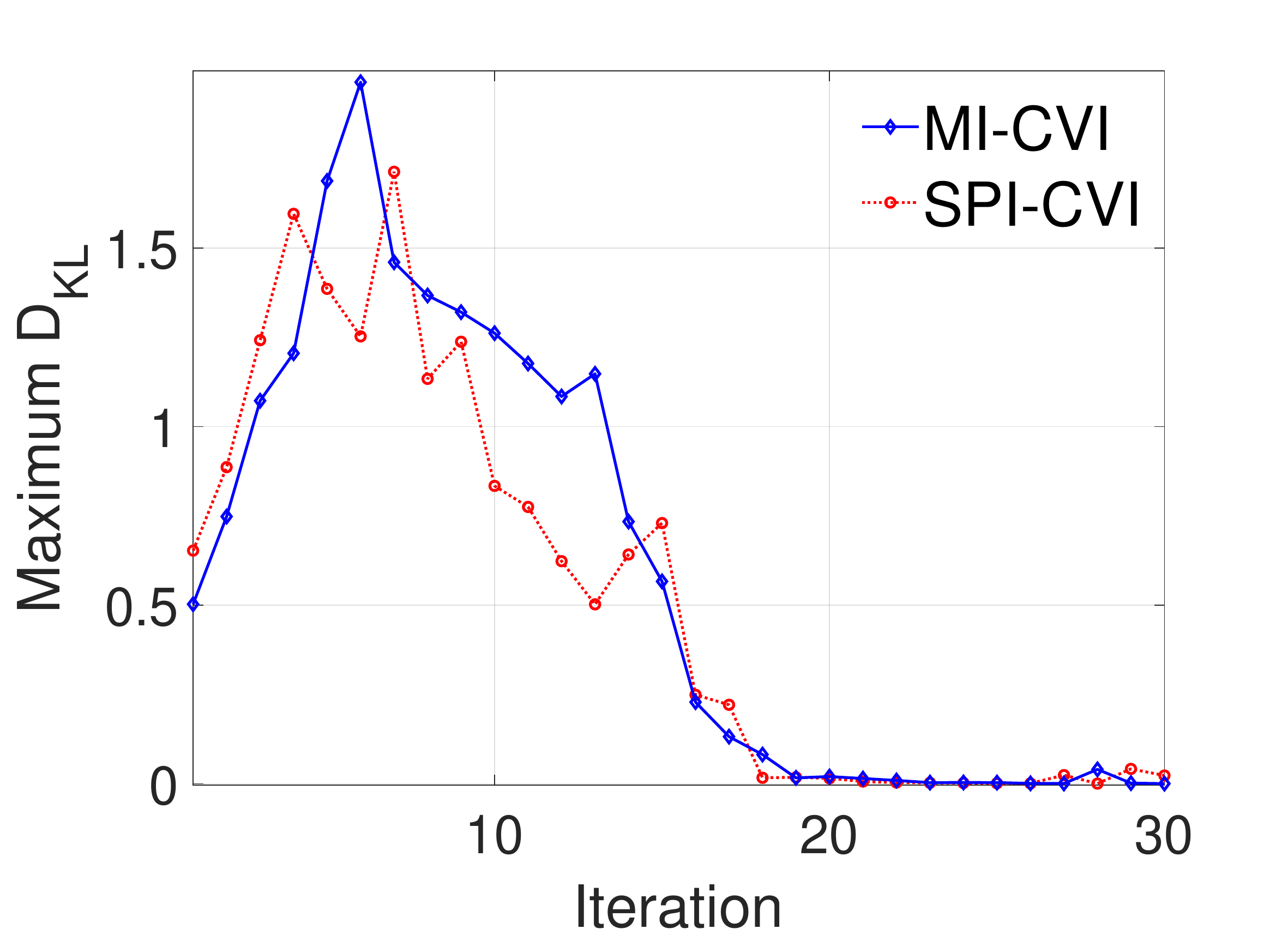}
    \caption{Maximum KL divergence}
    \label{policy_index}
  \end{subfigure}
  \caption{The gridworld environment (\ref{gridworld}) and comparison between SPI-CVI, MI-CVI and CVI. Black line shows mean SPI-CVI cumulative reward, blue line MI-CVI and red line CVI in (\ref{comp_cr}), with shaded area indicating $\pm 1$ standard deviation. Entering red rectangles incurs $-1$ reward, causing CVI converged slower. SPI-CVI is the upper bound of both MI-CVI and CVI. (\ref{gridworld_oscillation}) compares the respective policy oscillation value defined in Eq. (\ref{oscillation_measure}). (\ref{policy_index}) shows the maximum KL divergence $\max_{s} D_{KL}$ between policies of update.}
  \label{gridworld-rci}
\end{figure*}

\subsubsection{Update Rejection}

We discuss the optional rejection mechanism when $A^{{\pi_{k+1}}}_{{\pi_{k}},d} \!<\! 0$. When the computations are exact, greedy policy improvement always guarantee that $A^{{\pi_{k+1}}}_{{\pi_{k}},d} \!\geq\! 0$ \cite{pirotta13}. On the other hand, in value-based RL, inevitable sampling error and approximation might corrupt this guarantee. Thus, we propose to \emph{reject} the current $\pi_{k+1}$ and recollect samples for update by perturbing the policy $\pi_{k}$ a little. 


\subsubsection{Summary on the Algorithm}

Our proposed algorithm, Entropy-regularized Value-based RL with Monotonic Improvement is summarized in Alg. \ref{ARPU_alg}. Line 1 executes the initialization. Line 2 begins the main learning loop and lines 3 to 6 collect an episode of samples by interacting with the environment. Line 7 updates the value functions in-place using the sample pool $D_{0:k}$, and the updated value functions are then exploited in line 8 to update the policy $\pi_{k+1}$.  Lines 9 to 11 estimate $d^{\pi_{k}}$ and $A^{\pi_{k+1}}_{\pi_{k}}$. Line 12 computes the expected policy advantage. Line 14 produces $\tilde{\pi}_{k+1}$ based on the condition $A^{{\pi_{k+1}}}_{{\pi_{k}},d} \!\geq\! 0$. Line 15 corresponds to the optional rejection mechanism. If the updated policy $\pi_{k+1}$ is rejected, the learning is rolled back for recollecting samples and re-evaluating the policy.

\section{Experimental Results}\label{S4}

The proposed algorithm can be applied to a variety of entropy-regularized algorithms. In this section, we utilize conservative value iteration (CVI) in \cite{kozunoCVI} for experiments. 
In our implementation, 
for the $k+1$-th update, the baseline policy $\bar{\pi}$ in Eq. (\ref{sys_DPPbellman}) is $\pi_{k}$.  
We abbreviate the proposed algorithm based on CVI as \emph{monotonically improving CVI} (MI-CVI), and compare it also with \emph{safe policy iteration CVI} (SPI-CVI) \cite{pirotta13} from Lemma 2. In small state spaces, SPI-CVI performance should upper-bound that of MI-CVI, while for larger spaces, this guarantee might fail due to inaccurate estimates resulting from insufficient samples. All three algorithms are examined in both discrete and continuous state spaces.

\subsection{Gridworld with danger states}\label{S4sub1}

For discrete state space task, a stochastic 2-D gridworld problem with negative reward regions is solved using both CVI and MI-CVI. 
\subsubsection{Experimental Setting}\label{S4sub1s1}

 The agent in the gridworld shown in Fig. (\ref{gridworld}) starts from a fixed position and can move to any of its neighboring states with success probability $p$, or to a random different direction w.p. $1-p$. Its objective is to travel to a fixed destination and receive a $+1$ reward upon arrival. Stepping into red rectangles incurs a cost of $-1$. Every step costs $-0.1$ to encourage reaching the goal quickly. We maintain tables for value functions to inspect the case when there is no approximation error. Parameters are tuned to yield empirically best performance. For testing the \emph{sample efficiency}, every iteration terminates after 20 steps or upon reaching the goal, and only 30 iterations are allowed for training. For statistical significance the results are averaged over 100 independent trials. 
 
\subsubsection{Results}\label{S4sub1s2}

Fig. (\ref{comp_cr}) shows the performance of SPI-CVI, MI-CVI and CVI, respectively. Black, blue and red lines indicate their respective cumulative reward ($y$-axis) along the number of iterations ($x$-axis). Shaded area shows $\pm 1$ standard deviation. CVI learns policies that hit red rectangles more often and results in delayed convergence compared to the case with MI-CVI. 

Fig. (\ref{gridworld_oscillation}) compares the average \emph{policy improvement oscillation value} defined as:
\begin{align}
 \begin{split}
    &\forall k, \text{ such that } R_{k+1} - R_{k} < 0,\\
    &||\mathcal{O}J||_{\infty} = \max_{k} |R_{k+1} - R_{k}|, \\
    &||\mathcal{O}J||_{2} =  \sqrt{\big(\sum_{k} (R_{k+1} - R_{k})^{2}\big)}, 
 \end{split}
    \label{oscillation_measure}
\end{align}
where $R_{k+1}$ refers to the cumulative reward at $k+1$-th iteration. It is worth noting the difference $R_{k+1} - R_{k}$ is obtained by $\tilde{\pi}_{k+1}, \tilde{\pi}_{k}$, which is the lower bound of that by $\tilde{\pi}_{k+1}, {\pi}_{k}$. Intuitively, $|| \mathcal{O}J ||_{\infty}$ and $|| \mathcal{O}J ||_{2}$ measure \emph{maximum} and \emph{average} oscillation in cumulative reward. The stars between MI-CVI and CVI represent statistical significance at level $p=0.05$. In this problem the maximization in Eq. (\ref{J_first_exact}) is tractable, hence SPI-CVI upper bounds both the cumulative reward and oscillation value of MI-CVI. However, the difference between SPI-CVI and MI-CVI oscillation value is insignificant, suggesting the proposed algorithm is equally effective for small-scale problems as with SPI. 


The similar behavior of SPI-CVI and MI-CVI can also be verified in Fig. (\ref{policy_index}) which illustrates maximum KL divergence. Both algorithms show an increased divergence during the initial iterations, corresponding to discovery of better policies; and peak at around the middle stage of learning, then steadily decrease to zero, corresponding to convergence. This trend is similar to the $\Delta J^{\tilde{\pi}_{k+1}}_{\pi_{k}}$. The figures suggest that MI-CVI is capable of tightly approximating SPI-CVI behavior but with a much simplified computational procedure. In larger state spaces, the simplification is crucial as the proposed algorithm does not require estimating Eq. (\ref{J_first_exact}) accurately, which requires samples grow exponentially with dimensionality, as seen in the next example.

\begin{figure*}[t!]
  \begin{subfigure}[]{0.24\textwidth}
    \includegraphics[width=\linewidth]{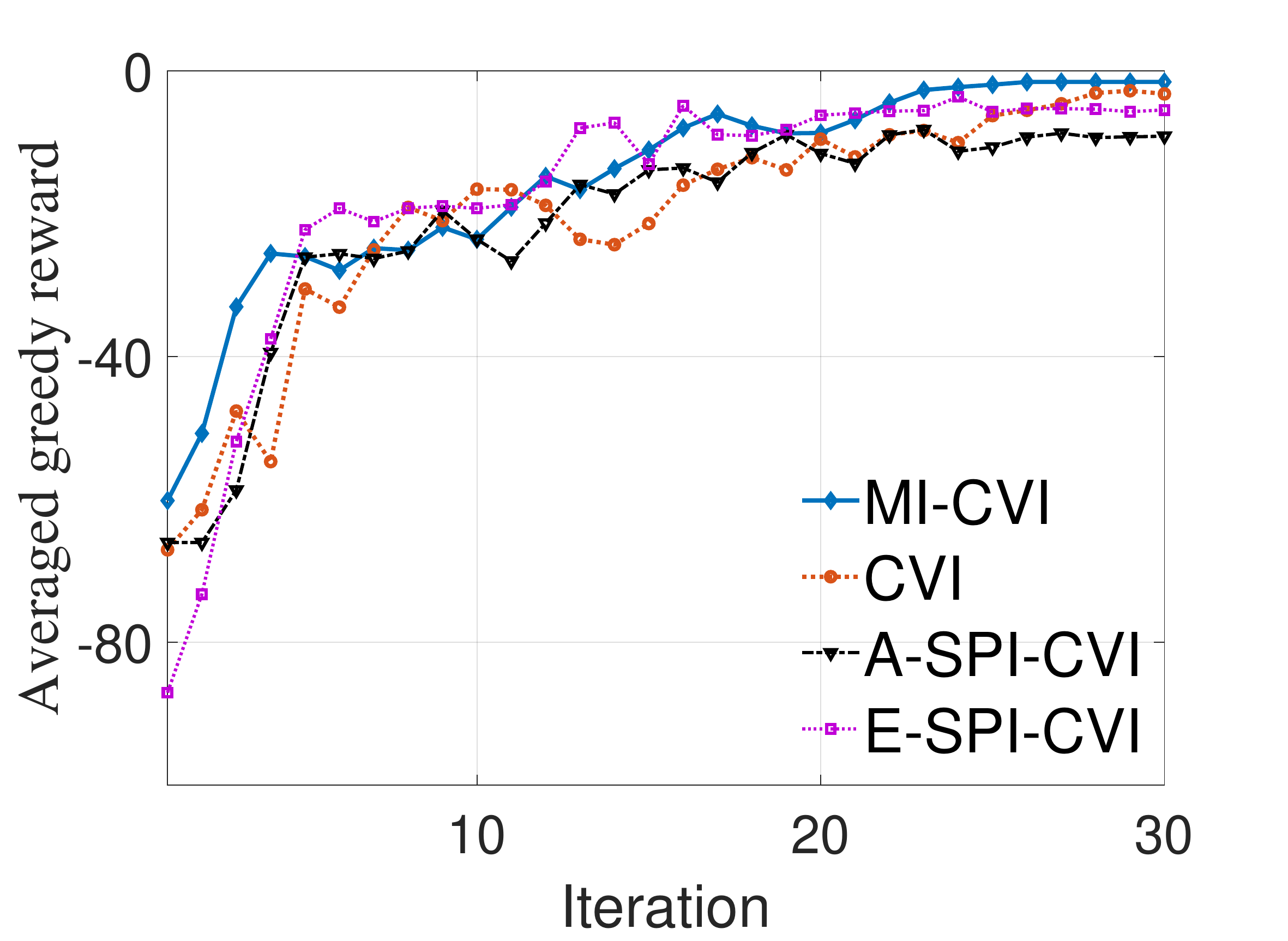}
    \caption{Cumulative reward}
    \label{pendulum_reward}
  \end{subfigure}
  \begin{subfigure}[]{0.24\textwidth}
    \includegraphics[width=\linewidth]{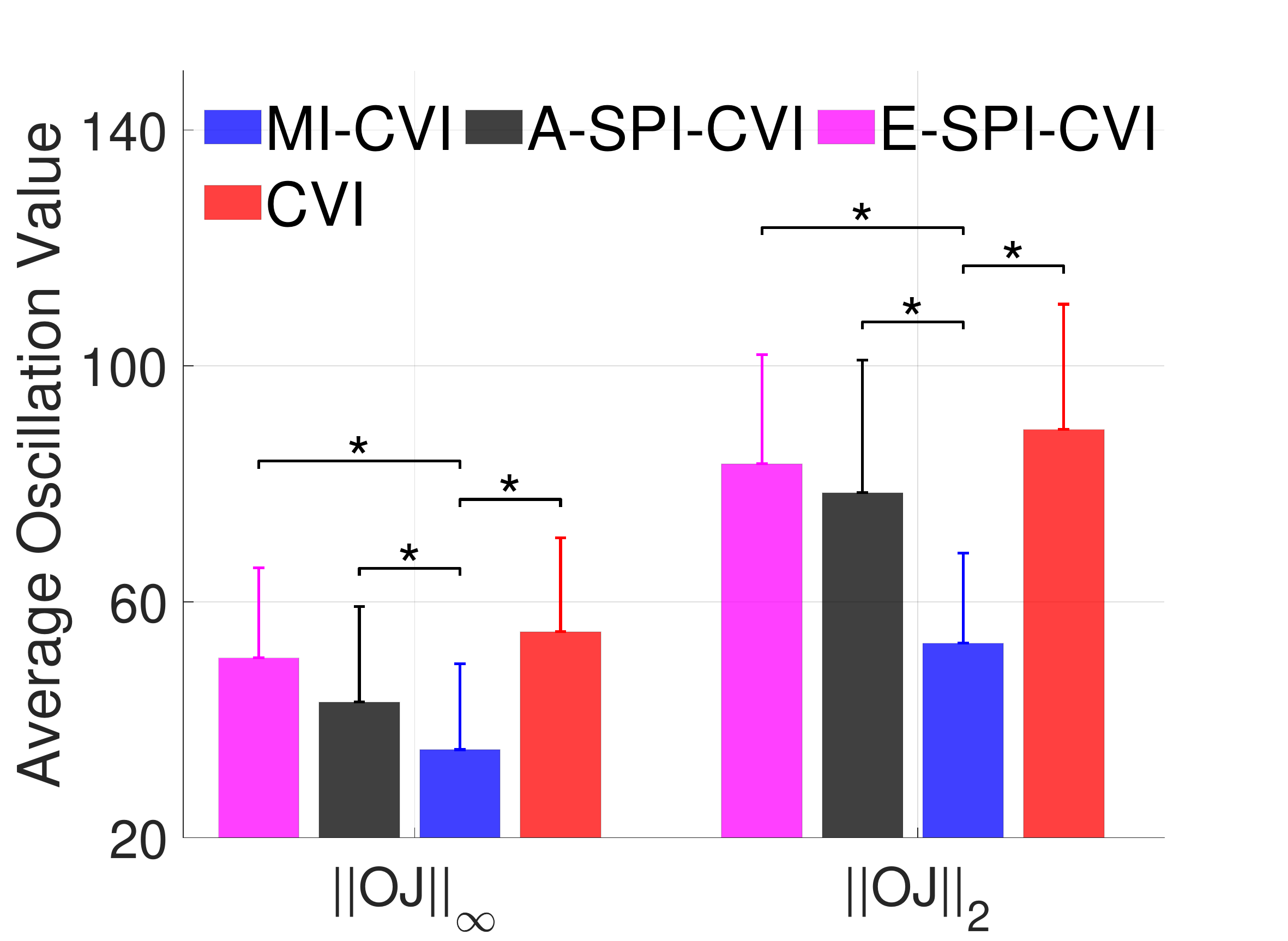}
    \caption{Values of policy oscillation}
    \label{pendulum_oscillation}
  \end{subfigure}
    \begin{subfigure}[]{0.24\textwidth}
    \includegraphics[width=\linewidth]{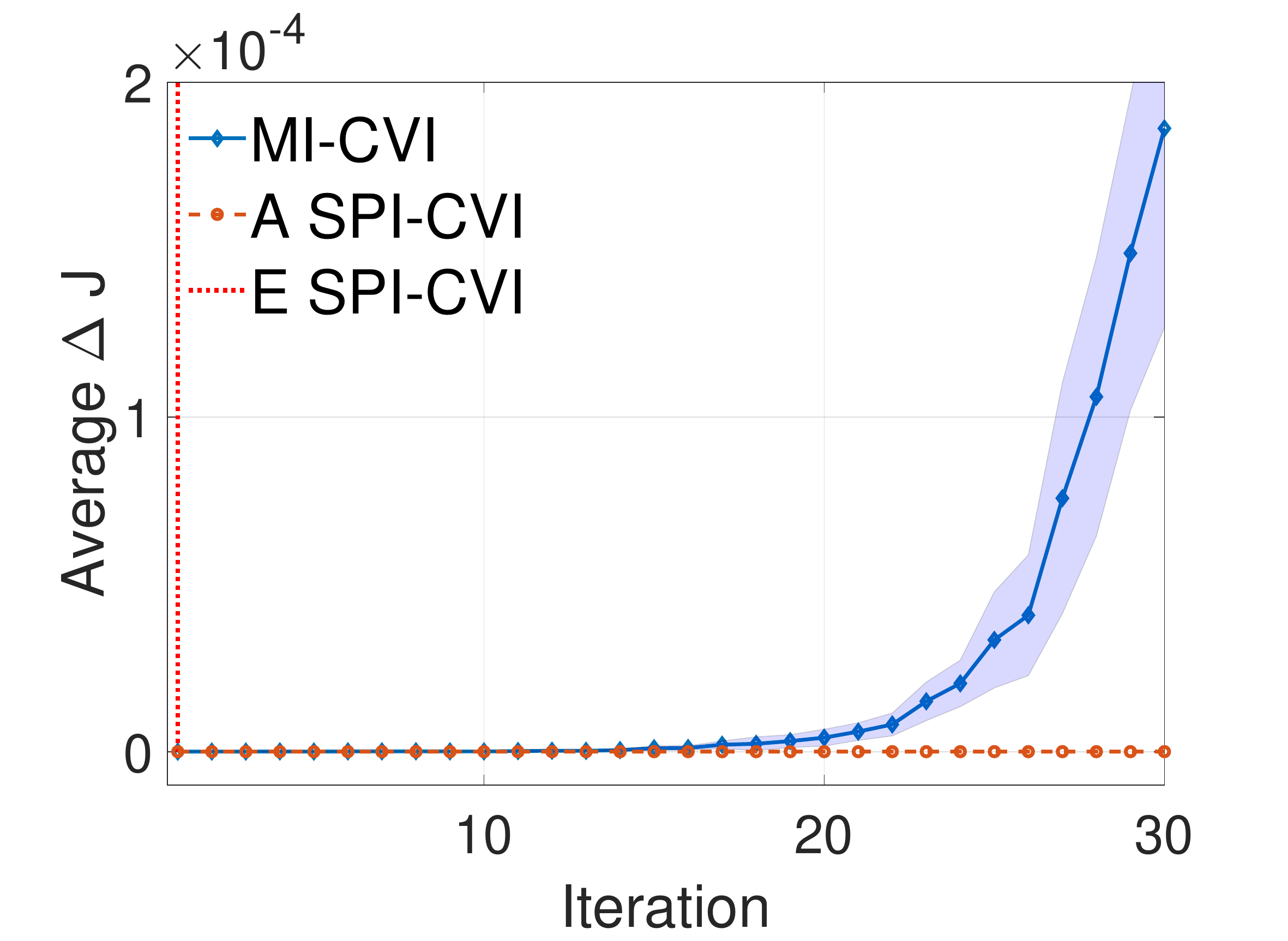}
    \caption{Values of $\Delta J^{\tilde{\pi}_{k+1}}_{\pi_{k}, d}$}
    \label{pendulum_imp}
  \end{subfigure}
  \begin{subfigure}[]{0.24\textwidth}
    \includegraphics[width=\linewidth]{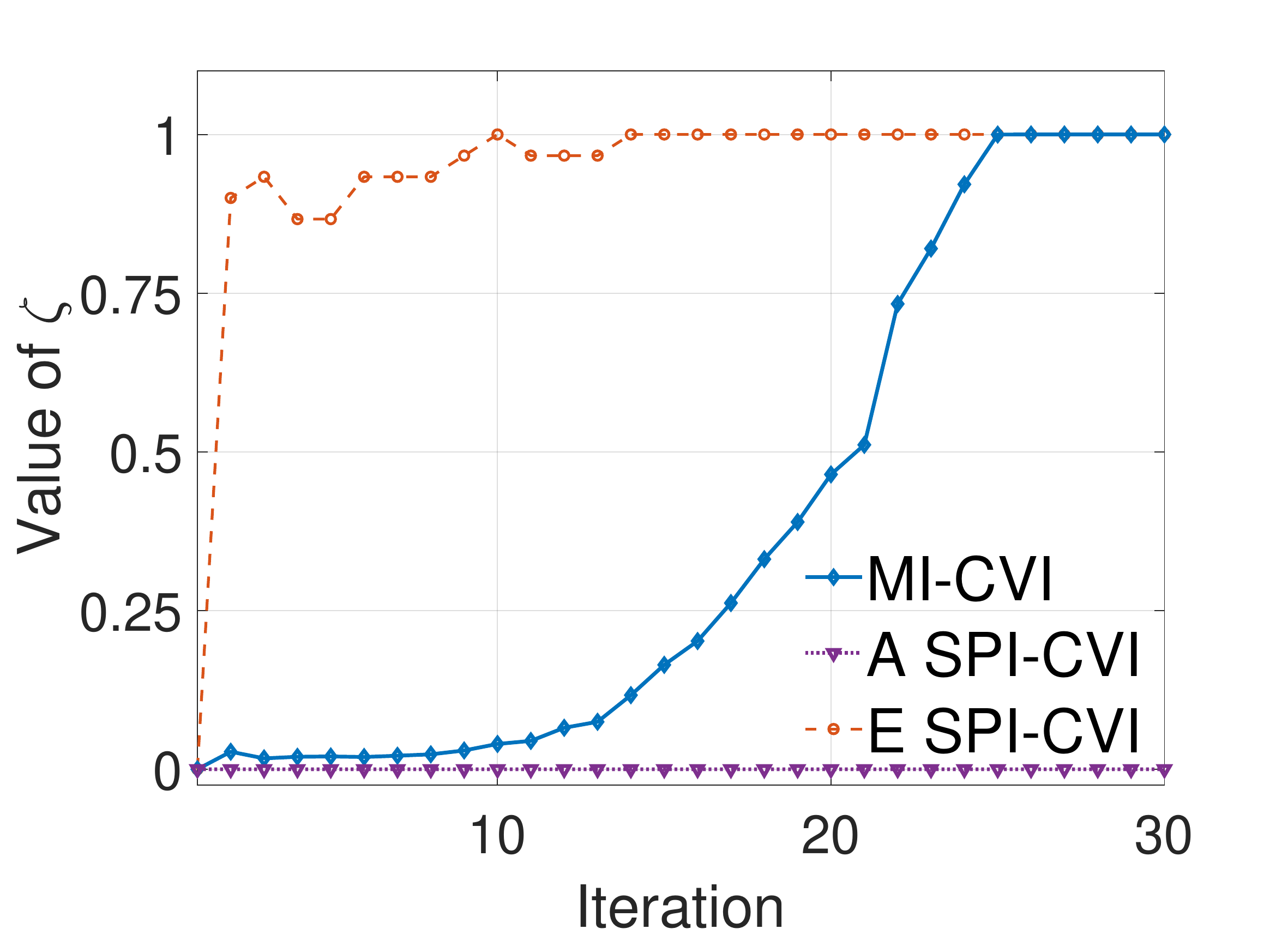}
    \caption{Values of $\zeta$}
    \label{pendulum_zeta}
  \end{subfigure}
  \caption{Comparison of SPI-CVI, MI-CVI and CVI on the pendulum swing-up task. In (\ref{pendulum_reward}) black, blue and red lines show the mean cumulative reward of SPI-CVI, MI-CVI and CVI, respectively. 
  (\ref{pendulum_oscillation}) illustrates policy oscillation value defined in Eq. (\ref{oscillation_measure}).
  (\ref{pendulum_imp}) shows the guaranteed improvement $\Delta J^{\tilde{\pi}_{k+1}}_{\pi_{k}, d}$ of MI-CVI, approximate SPI-CVI (A SPI-CVI) and exact SPI-CVI (E SPI-CVI). 
  (\ref{pendulum_zeta}) compares the values of $\zeta$ of MI-CVI and two versions of SPI-CVI.}
  \label{pendulum_results}
\end{figure*}

\subsection{Pendulum Swing-up}

In this section we examine all CVI-based algorithms on simulated pendulum swing-up, a classical control problem with continuous state space. Since direct implementation of SPI is not tractable, sampling-based SPI is employed. However, insufficient samples often lead to poor estimation of $\delta, \Delta$ and subsequently poor performance of SPI. One might instead wonder whether the relaxed version of Lemma 2 can be used, i.e., by using $\delta \Delta A^{\tilde{\pi}}_{\pi} \leq \frac{4}{1-\gamma}$ \cite{pirotta13}. We hence refer to the previously denoted SPI-CVI as \emph{E-SPI-CVI} (exact) and the new tractable version as \emph{A-SPI-CVI} (approximate) and compare them with MI-CVI.

\subsubsection{Experimental Setting}

A pendulum of length $1.5$ meters has a ball of mass $1$ kg at its end located at the fixed initial state $[0, -\pi]$. The pendulum attempts to reach the goal $[0, \pi]$ and stay there as long as possible. The state space is two-dimensional $s=[\theta, \dot{\theta} ]$, where $\theta$ denotes the vertical angle and $\dot{\theta}$ angular velocity. Action is one-dimensional torque $[-2, 0, 2]$ applied to the pendulum. The reward is defined to be negative quadratic in both the angle to the goal and angular velocity:
\begin{align*}
    R = - \frac{1}{z}(a\theta^{2} - b\dot{\theta}^{2}),
\end{align*}
where $\frac{1}{z}$ normalizes the rewards and large $b$ penalizes high angular velocity. We set $z = 10, a=1, b=0.01$.

Since the state space is continuous, function approximation has to be employed. We adopt linear function approximation (LFA) to approximate the Q-function by $Q(s,a) = \phi(s,a)^{T}\theta$, where $\phi(x) \!=\! [\varphi_{1}(x), \ldots, \varphi_{M}(x)]^{T}, x=[s,a]^{T}$, $\varphi(x)$ is basis function and $\theta$ corresponds to the weight vector. One typical choice of basis function is the radial basis function:
\begin{align*}
    \varphi_{i}(x) = \exp\big(-\frac{||x-c_{i}||^{2}}{\sigma^{2}}\big),
\end{align*}
where $c$ is the center and $\sigma$ is the width. We construct $\Phi = [\phi_{1}(x_{1}), \ldots, \phi_{M}(x_{N})] \in \mathbb{R}^{N\times M}$. For fast evaluation, the random features technique \cite{rahimi2008random} is used where $M \!=\! 800$ for all algorithms. To obtain the best-fit $\theta_{k+1}$ for $k \!+\! 1$-th iteration, the least-squares problem $||\mathcal{B}_{\pi}Q_{\pi_{k}} - \Phi\theta_{k} ||^{2}$ is solved:
\begin{align*}
    \theta_{k+1} = \big(\Phi^{T}\Phi + \alpha I\big)^{-1}\Phi^{T}\mathcal{B}_{\pi}Q_{\pi_{k}},
\end{align*}
where $\alpha$ is a small constant preventing singular matrix inversion, $\mathcal{B}_{\pi}Q_{\pi_{k}}$ is the empirical Bellman operator defined in Eq. (\ref{empirical_bellman}).

To demonstrate that the proposed algorithm can ensure monotonic improvement even with small number of samples, we allow 30 iterations of learning, each iteration comprises 200 steps. For statistical evidence, all figures show results averaged over 100 independent experiments.

\subsubsection{Results}

We compare MI-CVI with CVI and both approximate and exact versions of SPI-CVI in Fig. (\ref{pendulum_results}).
In Fig. (\ref{pendulum_reward}) the black line shows the mean value of cumulative reward of A-SPI-CVI, purple line of E-SPI-CVI, blue line of MI-CVI and red line that of CVI. E-SPI-CVI and CVI both exhibit wild oscillation in their curves, resulting in large average oscillation values in Fig. (\ref{pendulum_oscillation}). A-SPI-CVI, while being overly conservative, achieves smaller $||\mathcal{O}J||_{\infty}$, but not the $||\mathcal{O}J||_{2}$. On the other hand, MI-CVI learns smoothly thanks to the smooth growth of $\zeta$ from 0 to 1 and hence has significant less oscillation value than that of both SPI-CVI and CVI. The stars between MI-CVI and CVI, MI-CVI and both versions of SPI-CVI represent statistical significance at level $p=0.05$. 

The drastic behavior of SPI comes from the huge gap between exact and approximate SPI-CVI: in the E (exact) version, where insufficient samples lead to extremely small values of $\delta$ and $\Delta A^{\pi_{k+1}}_{\pi_{k}}$ and hence very large $\Delta J^{\tilde{\pi}_{k+1}}_{\pi_{k}, d}, \zeta$, as can be seen from Figs. (\ref{pendulum_imp}), (\ref{pendulum_zeta}). The aggressive choice of $\zeta$ leads to large oscillation value. On the other hand, A-SPI-CVI (approximate) takes the other extreme of producing vanishing $\zeta$ due to the loose bound $\delta \Delta A^{\tilde{\pi}}_{\pi} \leq \frac{4}{1-\gamma}$, as is obvious from the almost horizontal lines in the same figures: A-SPI-CVI has average value $\Delta J^{\tilde{\pi}_{k+1}}_{\pi_{k}, d} = 2.39\times 10^{-9}$ and $\zeta = 1.69\times 10^{-6}$.

By contrast, the advantage of MI-CVI is obvious: $\zeta$ can be tuned ranging from $0$ to $1$. While starting conservatively with $\zeta = 0$, MI-CVI is capable of leveraging the minor growth in advantage function (of magnitude $1\times 10^{-4}$) to update $\zeta$ gradually to 1, which corresponds to convergence.


\section{Discussion and Conclusion}\label{S5}

We proposed a novel lower bound on policy improvement for entropy-regularized value-based algorithms. Based on this, a novel RL algorithms was proposed to tackle the policy oscillation problem by ensuring monotonic policy improvement. The algorithm has been verified to ensure monotonic improvement in experiments with both discrete and continuous state spaces. In the latter, comparison with SPI demonstrates that the proposed algorithm is especially suitable for large state spaces.

Our future work includes applying the proposed algorithms on problems with higher dimensional state spaces with nonlinear function approximators such as deep networks \cite{Vieillard-2020DCPI}. For this, several theoretical points require further consideration such as changing the on-policy nature of the CPI and SPI to off-policy to fully leverage the merits of deep RL such as an experience replay technique.

Another interesting direction is to extend the current interpolation scheme from consecutive policies to any policies in a sequence. To this end, a number of technical difficulties should be addressed, e.g., proving that the Lemma 3 still applies to the resultant policy produced by interpolating several policies in a sequence.


\clearpage

\bibliography{library}
\bibliographystyle{aaai}

\clearpage

\appendix\label{apdx}

\subsection{Proof of Lemma 2}\label{crl3}

The proof was originally given by Pirotta et al. \cite{pirotta13}. For the ease of understanding Theorem 4 we rephrase it here. We also show that the role of $\zeta$ and $(1-\zeta)$ in Eq. (\ref{mixture_policy}) can be exchanged by solving a similar problem.

From Theorem 3.5 of \cite{pirotta13} we have:
\begin{align}
  \begin{split}
    \Delta J^{\pi'}_{\pi, d} &\geq A_{\pi,d}^{{\pi'}} - \frac{\gamma\Delta A_{\pi}^{{\pi'}}}{2(1-\gamma)^{2}}\max_{s}{\big|\!\sum_{a\in\mathcal{A}}\big({\pi'}(a|s)-\pi(a|s)\big)\big|}.
  \end{split}
  \label{pirotta}
\end{align}
Substituting in $\pi' = \zeta \tilde{\pi} + (1-\zeta)\pi$ one has:
\begin{align}
  \begin{split}
    A^{\pi'}_{\pi,d}&\!=\!\sum_{s}{\!d^{\pi'}{(s)}\!\sum_{a}{\pi'(a|s)A_{\pi}(s,a)}}\\
    &\!=\!\sum_{s}{\!d^{\pi'}{(s)}\!\sum_{a}{\big(\zeta\tilde{\pi}(a|s)+(1-\zeta)\pi(a|s)\big)A_{\pi}(s,a)}}\\
    & \!=\!\zeta\sum_{s}{\!d^{\pi'}{(s)}\!\sum_{a}{\tilde{\pi}(a|s)A_{\pi}(s,a)}}=\zeta A^{\tilde{\pi}}_{\pi,d},\\
    \Delta A^{{\pi'}}_{\pi}&\!=\!\max_{s, s'}{|A^{{\pi'}}_{\pi}(s)-A^{{\pi'}}_{\pi}(s')}|\\
    &\!=\! \max_{s, s'}{|\zeta A^{\tilde{\pi}}_{\pi}(s) -\zeta A^{\tilde{\pi}}_{\pi}(s')}|,\\
    \delta &= \max_{s}{\big|\!\sum_{a\in\mathcal{A}}\big({\pi'}(a|s)-\pi(a|s)\big)\big|},\\
    &= \max_{s}{\big|\!\sum_{a\in\mathcal{A}}\big(\zeta\tilde{\pi}(a|s)-\zeta\pi(a|s)\big)\big|}.\\
  \end{split}
  \label{SPI_loosen}
\end{align}
Hence Eq. (\ref{pirotta}) is transformed into:
\begin{align}
  \begin{split}
    \Delta J^{\pi'}_{\pi, d} &\geq \zeta A_{\pi,d}^{{\tilde{\pi}}} - \frac{\gamma\zeta^{2}\Delta A_{\pi}^{{\tilde{\pi}}}}{2(1-\gamma)^{2}}\max_{s}{\big|\!\sum_{a\in\mathcal{A}}\big({\tilde{\pi}}(a|s)-\pi(a|s)\big)\big|},
  \end{split}
  \label{SPI_quadratic}
\end{align}
the right hand side is a quadratic function in $\zeta$ and has its maximum at 
\begin{align}
  \begin{split}
\zeta^{*} = \frac{(1-\gamma)^2 A_{\pi,d}^{{\tilde{\pi}}}}{\gamma\Delta A_{\pi}^{{\tilde{\pi}}}\max_{s}{\big|\!\sum_{a\in\mathcal{A}}\big(\tilde{\pi}(a|s)-\pi(a|s)\big)\big|}}.
  \end{split}
\end{align}
By substituting $\zeta^{*}$ back to Eq. (\ref{SPI_quadratic}) we obtain that 
\begin{align}
  \begin{split}
    \Delta J^{\pi'}_{\pi, d} \geq \frac{\big((1-\gamma)A_{\pi,d}^{\tilde{\pi}}\big)^{2}}{2\gamma\delta\Delta A^{\tilde{\pi}}_{\pi}}.
  \end{split}
  \label{SPI_maximum}
\end{align}
In the case that $\zeta^{*}>1$, we clip it using $\min(1, \zeta^{*})$. 

Note that if we exchange the roles of $\zeta$ and $(1-\zeta)$, the coefficients in Eq. (\ref{SPI_loosen}) should be $(1-\zeta)$. Eq. (\ref{SPI_quadratic}) would become a quadratic function in $(1-\zeta)$, hence the r.h.s. of Eq. (\ref{SPI_maximum}) would be the maximum of $(1-\zeta^{*})$.
This concludes the proof.

\subsection{Proof of Theorem 4}\label{thm4}
\begin{proof}

We prove Theorem 4 by loosening $\delta$ and $\Delta A^{\tilde{\pi}}_{\pi}$ of Eq. (\ref{J_first_exact}):
\begin{align}
  \begin{split}
    &\Delta A^{\tilde{\pi}}_{\pi} = \max_{s, s'}{|A^{\tilde{\pi}}_{\pi}(s)-A^{\tilde{\pi}}_{\pi}(s')}|\\
    & \leq 2\max_{s}{|A^{\tilde{\pi}}_{\pi}(s)|}=2\max_{s}{\big|\!\sum_{a}{\!\tilde{\pi}(a|s)\! \big(Q_{\pi}(s,a)} \!-\! V_{\pi}(s)\big)\!\big|}\\
    & = 2\max_{s}{\!\big|\!\sum_{a}{\big(\tilde{\pi}(a|s)Q_{\pi}(s,a)-\pi(a|s)Q_{\pi}(s,a)\big)}\!\big|} \\
    & \leq 2 \max_{s}{\sum_{a}{\big|\big(\tilde{\pi}(a|s)-\pi(a|s)\big)Q_{\pi}(s,a)\big|}}\\
    &\leq  2\big|\!\big|Q_{\pi}(s,a)\big|\!\big|_{\infty}\max_{s}{\sum_{a}{\!\big|\tilde{\pi}(a|s)-\pi(a|s)\big|}} \\
     & \leq 2\sqrt{2}V_{max} \max_{s}{\sqrt{D_{KL}\big(\tilde{\pi}(\cdot|s)||\pi(\cdot|s)\big)}},
  \end{split}
  \label{J_exact_improved}
\end{align}
 where the second inequality makes use of the triangle inequality:
\begin{eqnarray}
  \begin{aligned}
    &\delta \leq \max_{s}\sum_{a\in\mathcal{A}}\big|\big(\tilde{\pi}(a|s)-\pi(a|s)\big)\big|,
    \label{triangle}
  \end{aligned}
\end{eqnarray}
and the third inequality makes use of Hölder's inequality $\frac{1}{p}\!+\!\frac{1}{q}\!=\!1$, with $p$ set to $1$ and $q$ set to $\infty$. The last inequality is because of Pinsker's inequality:
\begin{eqnarray}
\begin{aligned}
  &\max_{s}\sum_{a\in\mathcal{A}}\!\big|\tilde{\pi}(a|s)-\pi(a|s)\big|\!\leq \!\max_{s}{\!\sqrt{2D_{{KL}}(\tilde{\pi}(\cdot|s)||\pi(\cdot|s))}},
  \label{pinsker}
\end{aligned}
\end{eqnarray}
and the fact that $|\!|Q_{\pi}|\!|_{\infty}\leq V_{max}=\frac{1}{1-\gamma}$. By using the triangle inequality Eq. (\ref{triangle}) and Pinsker's inequality Eq. (\ref{pinsker}) we have:

\begin{align}
  \begin{split}
    &\Delta J^{\pi'}_{\pi, d} \geq \frac{\big((1-\gamma)A_{\pi,d}^{\tilde{\pi}}\big)^{2}}{2\gamma\delta\Delta A^{\tilde{\pi}}_{\pi}} \geq \frac{\big((1-\gamma)A_{\pi,d}^{\tilde{\pi}}\big)^{2}}{8\gamma \eta V_{max} }, \\
    &\delta=\max_{s}{\big|\!\sum_{a\in\mathcal{A}}\big(\tilde{\pi}(a|s)-\pi(a|s)\big)\big|},\\
    &\Delta A^{\tilde{\pi}}_{\pi}=\max_{s, s'}{|A^{\tilde{\pi}}_{\pi}(s)-A^{\tilde{\pi}}_{\pi}(s')}|,\\
    &\eta = \max_{s}{D_{KL}(\tilde{\pi}(\cdot|s)||\pi(\cdot|s))}.\\
  \end{split}
  \label{proof_leq}
\end{align}
By noting that the bound of Eq. (\ref{proof_leq}) can be loosened using Eq. (\ref{CVI_kl}), we obtain our pratical algorithm depending only on the tunable parameters $\alpha, \beta$ and $\gamma$:
\begin{align}
  \begin{split}
    &\eta\leq 2\beta\sum_{k=0}^{K-1}{\alpha^{k}\gamma^{K-k-1}},\\
    &\Delta J^{\pi'}_{\pi, d} \geq \frac{\big((1-\gamma)A_{\pi,d}^{\tilde{\pi}}\big)^{2}}{8\gamma \eta V_{max} } \geq \frac{\big(1-\gamma)^{3}(A_{\pi,d}^{\tilde{\pi}})^{2}}{16\gamma C_{K}},\\
    \text{where }C_{K} &= \beta\sum_{k=0}^{K-1}{\alpha^{k}\gamma^{K-k-1}}.\\
  \end{split}
\end{align}
Then the way of choosing $\zeta$ follows the proof in Lemma 2.

\end{proof}

\subsection{Proof of Lemma 5}
\begin{proof}

 For generality, we assume there is a initial state distribution $d_{0}$. For uncluttered notations, transition probability and policy are written in the matrix-vector form:
  \begin{align*}
    d^{\tilde{\pi}^{T}_{k+1}} &= d_{0}^{T} + \gamma d^{\tilde{\pi}^{T}_{k+1}} \mathcal{T}^{\tilde{\pi}_{k+1}},
  \end{align*}
where $d^{\pi} \in \mathcal{R}^{|\mathcal{S}|}$ is a vector, $\mathcal{T}^{\pi}=\Pi^{\pi}\mathcal{T}\in \mathbb{R}^{|\mathcal{S}|\times |\mathcal{S}|}$, $\mathcal{T} \in \mathbb{R}^{|\mathcal{S}||\mathcal{A}|\times|\mathcal{S}|}$, $\Pi^{\pi} \in \mathbb{R}^{|\mathcal{S}|\times|\mathcal{S}||\mathcal{A}|}$ are all stochastic matrices.

By noting that
\begin{align*}
  |\!| d^{\tilde{\pi}_{k+1}} - d^{\pi_{k}}  |\!|_{1} = |\!|  (d^{\tilde{\pi}_{k+1}} - d^{\pi_{k}})^{T}  |\!|_{\infty}
\end{align*} 
we can operate on the transpose of $d^{\tilde{\pi}_{k+1}} - d^{\pi_{k}}$. For uncluttered notations, we denote $\tilde{\pi}_{k+1}$ as $\alpha$ and $\pi_{k}$ as $\beta$: 
\begin{align*}
  & {{(d^{\alpha} - d^{\beta})^{T}}} = \gamma d^{\alpha^{T}}\mathcal{T}^{\alpha} - \gamma d^{\beta^{T}}\mathcal{T}^{\beta} \\
  &= \gamma {{\big( d^{\alpha} - d^{\beta} \big)^{T}}} \mathcal{T}^{\alpha} + \gamma d^{\beta^{T}}(\mathcal{T}^{\alpha} - \mathcal{T}^{\beta} )\\
  &= \gamma^{2} {{\big( d^{\alpha} - d^{\beta} \big)^{T}}} (\mathcal{T}^{\alpha})^{2} + \gamma d^{\beta^{T}}(\mathcal{T}^{\alpha} - \mathcal{T}^{\beta} ) \gamma\mathcal{T}^{\alpha} \\
  &= \gamma d^{\beta^{T}}(\mathcal{T}^{\alpha} - \mathcal{T}^{\beta} ) \sum_{t=0}^{\infty}\big(\gamma\mathcal{T}^{\alpha}\big)^{t}.
\end{align*}

Hence we see that 
\begin{align*}
  &|\!|  (d^{\alpha} - d^{\beta})^{T}  |\!|_{\infty} = |\!| \gamma d^{\beta^{T}}(\mathcal{T}^{\alpha} - \mathcal{T}^{\beta} ) \sum_{t=0}^{\infty}\big(\gamma\mathcal{T}^{\alpha}\big)^{t} |\!|_{\infty}\\
  & = \gamma |\!| d^{\beta^{T}} \! (\mathcal{T}^{\alpha} - \mathcal{T}^{\beta} ) (I - \gamma \mathcal{T}^{\alpha})^{-1} |\!|_{\infty} \\
  & \leq \gamma |\!| d^{\beta^{T}} |\!|_{\infty}  |\!| \mathcal{T}^{\alpha} - \mathcal{T}^{\beta} |\!|_{\infty} |\!| (I - \gamma \mathcal{T}^{\alpha})^{-1} |\!|_{\infty}\\
  & = \gamma  |\!| d^{\beta^{T}} |\!|_{\infty} |\!| \mathcal{T}(\Pi^{\alpha} - \Pi^{\beta}) |\!|_{\infty} |\!| (I - \gamma \mathcal{T}^{\alpha})^{-1} |\!|_{\infty}\\
  & \leq \frac{\gamma}{1 - \gamma}  |\!| \mathcal{T} |\!|_{\infty} |\!| \Pi^{\alpha} - \Pi^{\beta} |\!|_{\infty} |\!| (I - \gamma \mathcal{T}^{\alpha})^{-1} |\!|_{\infty} \\
  &\leq \frac{\gamma}{(1 - \gamma)^{2}} |\!| \Pi^{\alpha} - \Pi^{\beta} |\!|_{\infty}.\\
\end{align*}

Now we substitute back the definition of policies $\alpha$ and $\beta$:
\begin{align*}
  &\frac{\gamma}{(1 - \gamma)^{2}} |\!| \Pi^{\alpha} - \Pi^{\beta} |\!|_{\infty}\\
  & = \frac{\gamma}{(1 - \gamma)^{2}} \max_{s}\bigg|\! \sum_{a} (\tilde{\pi}_{k+1}(a|s) - \pi_{k}(a|s) \big)\bigg| \\
  &= \frac{\gamma}{(1 - \gamma)^{2}}  \zeta\max_{s}\bigg|\! \sum_{a} ({\pi}_{k+1}(a|s) - \pi_{k}(a|s) \big)\bigg|  \\
  &\leq  \frac{\zeta\gamma}{(1 - \gamma)^{2}} \sqrt{2 \max_{s}D_{KL}(\pi_{k+1} || \pi_{k})} \\
  &\leq \frac{2\zeta\gamma}{(1 - \gamma)^{2}} \sqrt{ C_{K}},
\end{align*}
where the penultimate inequality follows from the Pinsker's inequality and the last inequality follows from Lemma 3.

\end{proof}

\subsection{Proof of Theorem 6}
\begin{proof}
  Starting from the definition, we have: 
\begin{align*}
  &\big| \mathcal{L}^{\alpha, d^{\alpha}}_{\beta, d^{\beta}} \big| = \big|    (d^{\alpha} - d^{\beta})^{T} A_{\beta}^{\alpha}\big|\\  
  &\leq |\!| (d^{\alpha} - d^{\beta})^{T} |\!|_{\infty}  |\!|A_{\beta}^{\alpha}|\!|_{1}   \\
  &\leq  \frac{2\zeta\gamma}{(1-\gamma)^2} C_{K} |\!|A_{\beta}^{\alpha}|\!|_{1}  \\
  &\leq (1-\gamma) |\!|A_{\beta}^{\alpha}|\!|^{2}_{1}, 
\end{align*}
where the first inequality leverages Hölder's inequality and the second inequality makes use of Lemma 2. The last inequality follows from substituting in $\zeta = \frac{(1-\gamma)^{3}A^{{\pi_{k+1}}}_{{\pi_{k},d}}}{2\gamma C_{K}}$.
  
\end{proof}

\end{document}